\documentclass[10pt,journal]{IEEEtran}
\usepackage{amssymb}
\usepackage{algorithm, algorithmic}
\usepackage{graphicx}
\usepackage{epstopdf}
\usepackage{gensymb}
\usepackage{color}
\usepackage{amsmath}
\usepackage{bm}
\usepackage{etoolbox}
\usepackage{makecell}

\usepackage{cite}
\usepackage{array}
\usepackage{booktabs}
\usepackage{multirow}
\usepackage{comment}
\usepackage{subcaption}

\usepackage{mathrsfs}
\usepackage{hyperref}
\usepackage{wasysym}

\usepackage{balance}
\usepackage{soul}
\hypersetup{
    colorlinks=true,
    linkcolor=black,
    filecolor=black, 
    citecolor=black,
    urlcolor=cyan,
}

\usepackage[framemethod=TikZ]{mdframed}
\usepackage[all]{genealogytree}

\usepackage{threeparttable}
\usepackage{mathtools}
\usepackage{hyperref}
\usepackage{pifont}
\usepackage{listings}

\usepackage{url}
\urldef{\mailsa}\path|{alfred.hofmann, ursula.barth, ingrid.haas, frank.holzwarth,|
	\urldef{\mailsb}\path|anna.kramer, leonie.kunz, christine.reiss, nicole.sator,|
	\urldef{\mailsc}\path|erika.siebert-cole, peter.strasser, lncs}@springer.com|    

\usepackage[utf8]{inputenc}
\usepackage[english]{babel}

\usepackage{amsthm}
\usepackage{balance}

\usepackage{bbding}

\theoremstyle{definition}

\definecolor{R}{RGB}{0,0,150}

\theoremstyle{remark}

\ifCLASSINFOpdf

\definecolor{blue}{RGB}{60,132,196}
\definecolor{red}{RGB}{207,78,56}
\definecolor{gray}{RGB}{146,146,161}

\newcommand{\eat}[1]{}

\begin{document}

\title{Vertical Federated Learning: Taxonomies, Threats, and Prospects}

\author{
        Qun Li,
        Chandra Thapa,
        Lawrence Ong,
        Yifeng Zheng, Hua Ma,
        Seyit A. Camtepe,
        Anmin Fu,
        Yansong Gao\IEEEauthorrefmark{2}
	\thanks{Q.~Li, A.~Fu and Y.~Gao are with the School of Computer Science and Engineering, Nanjing University of Science and Technology, China.
	\{120106222757;yansong.gao;fuam\}@njust.edu.cn}
 
	\thanks{C.~Thapa and S.~Camtepe are with Data61, CSIRO, Australia. E-mail: \{chandra.thapa;seyit.camtepe\}@data61.csiro.au}
        \thanks{L.~Ong is with The University of Newcastle, Newcastle, NSW, Australia. E-mail:  lawrence.ong@newcastle.edu.au}
	\thanks{Y.~Zheng is with Harbin Institute of Technology, Shenzhen, China. E-mail: yifeng.zheng@hit.edu.cn}
        \thanks{H.~Ma is with the University of Adelaide, Adelaide, Australia and Data61, CSIRO, Australia. E-mail: hua.ma@adelaide.edu.au}
   \thanks{Y.~Gao\IEEEauthorrefmark{2} is the corresponding author.}
   
}

\maketitle

\begin{abstract}
Federated learning (FL) is the most popular distributed machine learning technique. 
FL allows machine-learning models to be trained without acquiring raw data to a single point for processing. Instead, local models are trained with local data; the models are then shared and combined. This approach preserves data privacy as locally trained models are shared instead of the raw data themselves. Broadly, FL can be divided into horizontal federated learning (HFL) and vertical federated learning (VFL). For the former, different parties hold different samples over the same set of features; for the latter, different parties hold different feature data belonging to the same set of samples. In a number of practical scenarios, VFL is more relevant than HFL as different companies (\emph{\emph{e.g.}}, bank and retailer) hold different features (\emph{\emph{e.g.}}, credit history and shopping history) for the same set of customers. Although VFL is an emerging area of research, it is not well-established compared to HFL. Besides, VFL-related studies are dispersed, and their connections are not intuitive. Thus, this survey aims to bring these VFL-related studies to one place. Firstly, we classify existing VFL structures and algorithms. Secondly, we present the threats from security and privacy perspectives to VFL. Thirdly, for the benefit of future researchers, we discussed the challenges and prospects of VFL in detail.

\end{abstract}

\IEEEpeerreviewmaketitle

\section{Introduction}\label{sec:Intro}

Deep learning (DL) has enabled a wide range of applications such as facial recognition, speech recognition, language translation, object detection, medical diagnosis, and gene expression~\cite{lecun2015deep,jumper2021highly}. One core driver of the excellent performance of the deep learning model is the availability of huge amounts of data.

However, due to the restriction enforced by the privacy protection regulations, such as the General Data Protection Regulation (GDPR, effective from May 2018)~\cite{gdpr}, California Privacy Rights Act (CPRA, effective from January 2021)~\cite{cpra}, and China Data Security Law (CDSL, effective from September 2021)~\cite{cdsl}, it is always a challenge to access or share the raw data directly. When it comes to sensitive data such as medical data (\emph{\emph{e.g.}}, cancer diagnosis) or bank records (\emph{\emph{e.g.}}, credit card usage), even the data custodians are highly reluctant to share these data. This results in data silos or islands. Moreover, data silos are the major cause of the unavailability of sufficient data for DL model training. So, these silos need to be broken, and collaborative model training approaches break these data silos in DL model training.

Google introduced federated learning (FL) in 2016 that enables collaborative DL model training among distributed participants without sharing the raw data held by the participants~\cite{konevcny2016federated}. In FL, data resides locally, and only intermediate gradient information of the local model is transmitted to the FL coordinator (\emph{i.e.}, server). 
This way, FL greatly alleviates the risk of data privacy leakage~\cite{gao2021evaluation} and thus complies with the regulations. 
Considering the commonalities either in the sample or feature spaces of the data distributed among the participants, FL is broadly divided into two categories, namely horizontal federated learning (HFL) and vertical federated learning (VFL).

\begin{figure}[t]
    \centering
    \includegraphics[trim=5cm 3cm 5cm 3.5cm, clip=true, width=0.8\columnwidth]{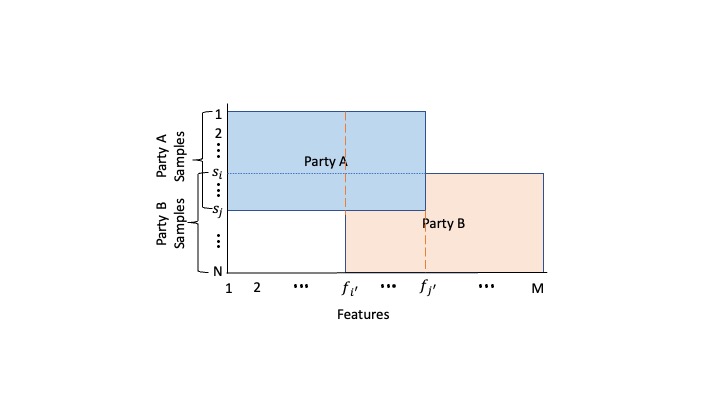}
     \caption{An illustration of sample space with $N$ total samples and feature space with $M$ total features in federated learning with two parties, Party A and Party B. The sample and feature sets for (i) HFL are $\{1,2,\dotsc,N\}$ and
$\{{f_{i^{'}}}, {f_{{i^{'}}+1}},\dotsc,{f_{j^{'}}}\}$, 
where ${{i^{'}} \leq {j^{'}}}$, and (ii) 
     VFL are $\{s_{i}, s_{i+1},\dotsc,s_{j}\}$, 
     where $ i\leq j$, and $\{1, 2,\dotsc,M\}$.}
    \label{fig:fl}
\end{figure}

HFL considers the distributed clients having data with the same features but are different in sample space~\cite{feng2021blockchain,zhao2021efficient}. 
For example, patient data in two hospitals in different regions.
In contrast to HFL, VFL considers the distributed clients having the same samples but different features~\cite{jin2021cafe,zhang2021secure}. 
The VFL is more often related to an enterprise setting with \textit{a smaller number} of participants compared to the HFL, but privacy is more paramount~\cite{kang2022framework}.
For example, data in a bank and a hospital from the same region. As the two entities are in the same region, their samples (users) can be common but different in feature space because these two entities have different businesses. Figure~\ref{fig:fl} gives an illustration of HFL and VFL with two parties. 
The sample space consists of $N$ total samples, and the feature space has $M$ total features within two parties, Party A and Party B. The sample and feature sets for (i) HFL are $\{1,2,\dotsc,N\}$ and $\{{f_{i^{'}}}, {f_{{i^{'}}+1}},\dotsc,{f_{j^{'}}}\}$, where ${{i^{'}} \leq {j^{'}}}$, and (ii) 
     VFL are $\{s_{i}, s_{i+1},\dotsc,s_{j}\}$, 
     where $ i\leq j$, and $\{1, 2,\dotsc,M\}$. We note that there is a third category of FL, called transfer federated learning, that considers no commonalities in the feature and sample space of the data among the participants. In other words, this is the category when $j< i$ and $j'<i'$ in Figure~\ref{fig:fl}. We exclude such cases in this study as it requires a different technique, called transfer learning, which enables DL by transferring information from one domain to a related domain~\cite{transferlearning}. 

\begin{table*}[]
    \centering
    \caption{Comparison between existing preliminary VFL-related surveys and our comprehensive survey covering four major aspects: categorization, privacy \& security, application, and challenges \& prospects.}
    \scalebox{0.95}{
    \begin{tabular}{c|c|c|c|c}
    \hline
       \textbf{Work} & \textbf{VFL categorization} & \makecell[c]{\textbf{Privacy \&  security issues}} & \textbf{Application scenarios} & \textbf{Challenges \& prospects}\\
    \hline
    Yang {\it et al.}~\cite{yang2019federated} & \XSolidBrush& \XSolidBrush& \Checkmark & \XSolidBrush\\
    \hline
    Wei {\it et al.}~\cite{wei2022vertical} & \XSolidBrush & \Checkmark & \XSolidBrush & \Checkmark\\
    \hline
    Xu {\it et al.}~\cite{xu2022privacy} & \XSolidBrush& \XSolidBrush& \XSolidBrush & \Checkmark\\
    \hline
    Our work & \Checkmark & \Checkmark & \Checkmark & \Checkmark \\
     \hline
     
    \end{tabular}
    }

    \label{tab:comapre1}
\end{table*}

Existing FL reviews mainly concentrate on HFL~\cite{bonawitz2019towards,li2020federated}, while VFL studies are less combed.
In fact, there are varying VFL algorithms in the literature with diverse customized assumptions, and yet, no investigation work has been dedicated to unifying them.
In addition, the privacy and security issues faced by the VFL are also less elucidated, and this may lead to misuse or false security implications. Overall, it is important to comb these diverse VFL algorithms and analyze the security and privacy threats, as well as corresponding countermeasures, 
so as to facilitate future research and ease the adoption of VFL.

The rest of the paper is organized as follows. Section~\ref{sec:related} discusses related works that preliminarily surveyed the VFL studies and distinguishes our survey from them. Section~\ref{sec:preliminary} provides necessary preliminaries. 
Section~\ref{sec:classification} elaborates on combing existing VFL algorithms to classify them, followed by applications per categorization. Security and privacy threats faced by the VFL are detailed in Section~\ref{sec:SecPri}. Challenges and Prospects are discussed and provided, respectively, in Section~\ref{sec:ChaPro}. This work is concluded in Section~\ref{sec:conclusion}.

\section{Related Works}\label{sec:related}

With increasing applications of VFL in various fields, few preliminary survey works related to VFL emerged. However, they often only take partial aspects (\emph{i.e.}, application scenarios, or faced challenges) into consideration. Yang {\it et al.}~\cite{yang2019federated} divided FL into three categories: HFL, VFL, and transfer federated learning (TFL). 
They then provided a definition, a general structure, and a potential application for each of these three FL types. However, this work does not take into account the possible (unique) privacy and security issues and corresponding challenges confronted by VFL. And it also does not detail different VFL algorithms. Wei {\it et al.}~\cite{wei2022vertical} studied the potential challenges and unique issues of VFL from four aspects: security and privacy risks, computing and communication overhead, structural destruction, and system heterogeneity. 
They, however, only focus on one type of VFL structure: the split neural network, so the important VFL taxonomy is missing. In addition, neither the privacy risk of attribute inference nor the  security threats, such as backdoor attacks, are elucidated. Xu {\it et al.}~\cite{xu2022privacy} reviewed several privacy-preserving VFL schemes and showed their differences in terms of communication, computing requirements, number of entities required, and trust assumptions. Although this work compared several privacy-preserving VFL schemes, it only introduced limited VFL schemes, neither extending nor classifying them.

As summarized in Table~\ref{tab:comapre1}, existing preliminary VFL-related surveys are not comprehensive and merely take partial aspects into account. Therefore, our work attempts to take a step forward to comprehensively review existing studies on VFL. 
First, we classify existing devised VFL structures and algorithms in a single framework. Secondly, we introduce the application scenarios of different types of VFL. Then, compared to~\cite{wei2022vertical} that merely mentions some privacy issues, such as the existence of label inference and data reconstruction but without elaborating on attacking methods, we comprehensively analyze the threats to VFL from the perspective of security and privacy. Finally, in order to benefit future research, we discuss the challenges and provision prospects of VFL.

\section{Preliminary}\label{sec:preliminary}

This section introduces VFL participants and the VFL conceptual definition. The decision tree is succinctly described as it is later used to customize a special VFL design.

\subsection{VFL Participants}

In VFL, there are typically three types of participants, namely active participants, passive participants, and a coordinator.

\noindent\textbf{Active participant:} An active participant is  an entity in VFL that wants to build models and provide data with a sample set $\{s_i,\cdots,s_j\}$ and their labels $\{y_i\cdots,y_j\}$.

\noindent\textbf{Passive participant:} A passive participant is an invited entity in VFL that provides data with the sample set $\{s_i,\cdots,s_j\}$ but with no labels, and its feature set is different than that of the active participant. There can be multiple passive participants in VFL.

\noindent\textbf{Coordinator:} A coordinator coordinates the training process in VFL, and it communicates with the participants in an encrypted channel. It has no access to the raw data. In real-world scenarios, the role of the coordinator is usually performed by the active participant or a trusted third party.

\subsection{VFL Definition}

VFL is a distributed machine learning technique that considers distributed data with the same sample space but different feature spaces among the participants. Suppose there are $P$ passive participants {$C_1,...,C_P$}, their local data are {$\mathcal{D}_1,...,\mathcal{D}_P$}, and one active participant $C_\textup{T}$ providing the label space $\mathcal{Y}$. All participants have their own local models. The model of a participant is represented as $\theta$. The optimization objective of VFL is expressed as:

\begin{equation}
    \min_{\theta_1,...,\theta_P,\theta_\textup{T}}\mathcal{L}(\theta_1,...,\theta_P,\theta_\textup{T};\mathcal{D},\mathcal{Y}),
\end{equation}
where $\mathcal{L}$($\cdot$) is the loss function of global model, $\theta_\textup{T}$ is the local model of the active participant and $\mathcal{D} = \mathcal{D}_1\cup  \cdots \cup\mathcal{D}_P\cup\mathcal{D}_\textup{T}$.

\subsection{VFL categories}
VFL can be categorized broadly into two types based on the data distribution among the participants:
\begin{itemize}
    \item \textbf{Heterogeneous vertical federated learning}: In this category, VFL participants (passive and active) have different data distributions. This adds up difficulties in machine learning convergence. Thus careful preprocessing of data, \emph{e.g.}, normalization and feature selection, is required. Usually, VFL with participants of different domains, for example, hospitals and banking, falls under heterogeneous VFL.

    \item \textbf{Homogeneous vertical federated learning:} In this category, VFL participants have similar data distributions. Usually, VFL with participants of similar domains, for example, hospitals and pathology, falls under homogeneous VFL.
\end{itemize}

\subsection{Decision Tree}
One class of VFL algorithms uses customized decision trees. To facilitate subsequent understanding, the decision tree is briefly introduced in the following.

Decision tree~\cite{myles2004introduction,song2015decision} is a typical machine learning model for decision-making with the tree structure, which can be used for classification or prediction. The decision tree model consists of many internal nodes and leaf nodes. Each internal node corresponds to an attribute. Each leaf node represents a class label. When predicting an input, a series of branching operations from the root node down to the leaf node are performed according to the comparison of the input's attribute value and the internal node's attribute threshold value until a leaf node is reached. The prediction output is the class label on the leaf node. 

The construction/training of the decision tree is as follows: build the root node, put all the training data on the root node, select an optimal feature or attribute, and divide the training dataset into subsets to ensure the training set has the best classification under the current split. If these subsets can be well classified, leaf nodes are constructed and ended. Otherwise, it continuously separates them and constructs internal nodes, and this process is repeated until all training sets are correctly classified with the desired accuracy or there are no features that can be used for splitting.


\section{VFL Algorithms}\label{sec:classification}

\begin{figure} [t]
    \centering
    \begin{center}
\begin{tikzpicture}
  \genealogytree[timeflow=left,template=signpost,box={height=3em,width=4.7em},
    edges={foreground={line width=1pt,red,LaTeX-,shorten <=-4pt},
    background={line width=1.5pt,yellow},
  hide single leg,},
  ]
  {
    
      parent{
        g{VFL \\Algorithms}
        parent
        {
          g{Non-split VFL}
          parent{
            g{Non-split VFL with coordinator~\cite{jin2021cafe,benmalek2022security,sun2021vertical} }
          }
          parent{
            g{Non-split VFL without coordinator~\cite{zhang2021secure,zhang2022data,liu2020asymmetrical}}
          }
        }
        parent{
          g{Split
          VFL~\cite{zhang2021desirable,hashemi2021vertical,kang2020fedmvt}}
        }
        parent{
          g{Customized VFL~\cite{wu2020privacy,luo2021feature,tian2020federboost}}
        }
      }
  }
\end{tikzpicture}
\end{center}
    \caption{Taxonomy of VFL algorithms.}
    \label{fig:vfl_categories}
\end{figure}
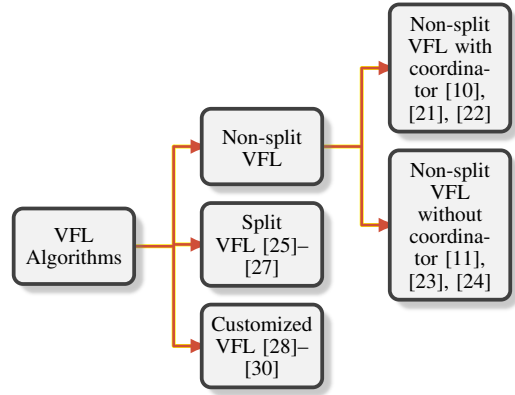

According to whether a partial model portion or the full model is provided to the participants,
we divide VFL into (i) non-split VFL, (ii) split VFL, and iii) customized VFL. In non-split VFL, each participant owns a full model, whereas the model is split, and each participant owns part of the model in the split VFL. We note there is a customized VFL based on a decision tree, which is not using a typical deep learning model (\emph{i.e.}, building upon cascading CNN or/and fully connected layers). We thus refer to these VFL algorithms as the customized VFL. The taxonomy of VFL algorithms is summarized in Figure~\ref{fig:vfl_categories}.

\subsection{Non-split VFL}

In the non-split VFL, each participant has a full model, and it computes the gradient according to (i) the local data it owns and (ii) the intermediate information (\emph{i.e.}, softmax) provisioned by other participants. According to the different organizing means of participants, the non-split VFL can be further divided into non-split VFL with coordinator~\cite{jin2021cafe,benmalek2022security,sun2021vertical} and non-split VFL without coordinator~\cite{zhang2021secure,zhang2022data,liu2020asymmetrical}.

\subsubsection{Non-split VFL with coordinator} 

\begin{figure}
    \centering
    \includegraphics[width=0.3\textwidth]{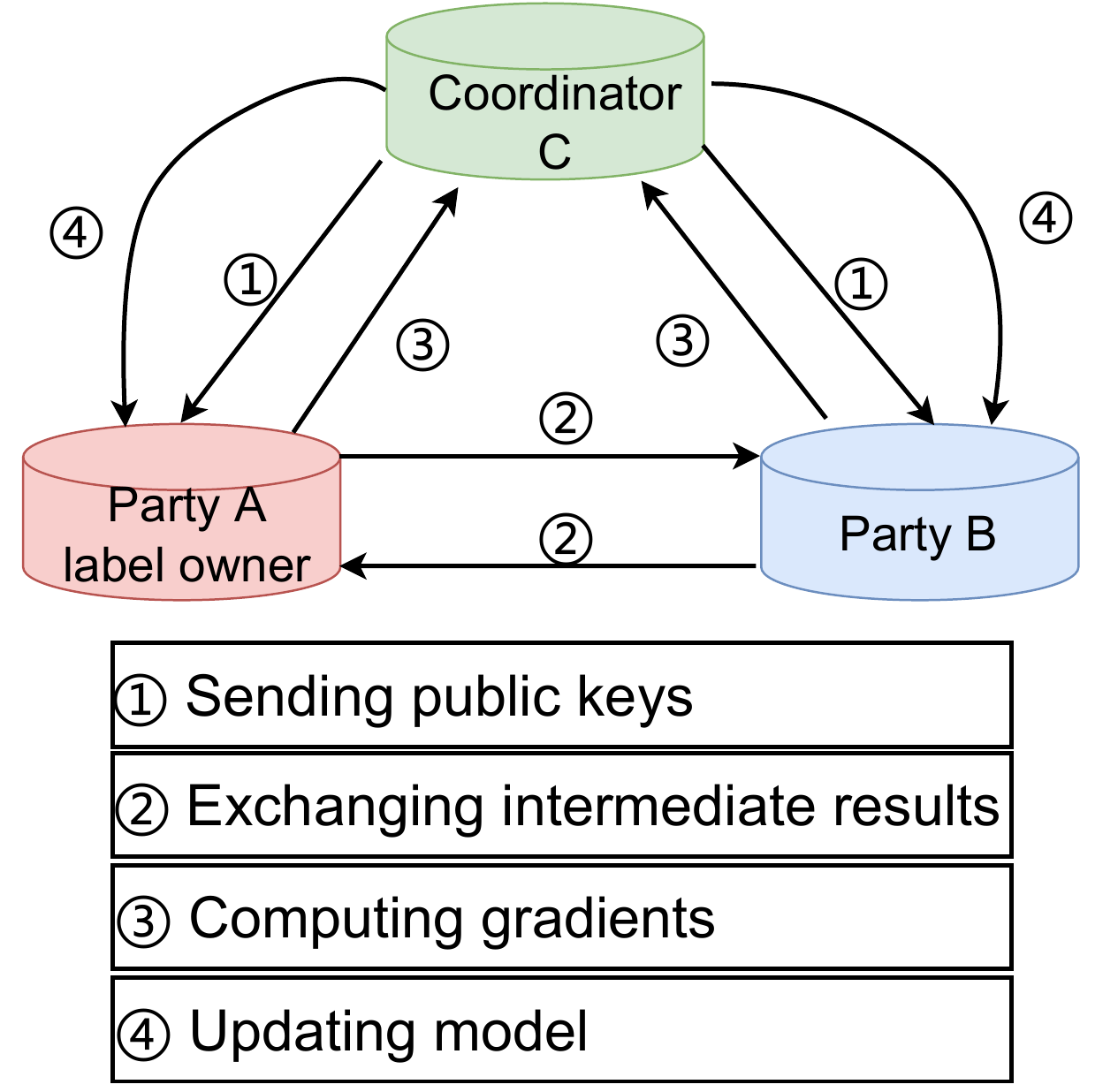}
    \caption{Non-split VFL with coordinator.}
    \label{fig:with coordinator}
\end{figure}

As shown in Figure~\ref{fig:with coordinator}, it has three types of participants, namely active participant Party A who owns the label; passive participant Party B; and coordinator C. The training process is divided into 5 steps. Algorithm \ref{nowith} shows the training process, where [[$\cdot$]] denotes ciphertext encrypted with the public key. The value $\mathcal{R}$ used for masking is a random number generated by each participant.

\noindent\textbf{Step \textcircled{1}.} The coordinator C generates a key pair ($k_p,k_s$).
The $k_p$ stands for the public key and $k_s$ stands for the private key. Then C sends the public key $k_p$ to A and B. A and B initialize (\emph{i.e.}, randomize the value of the weight) their local models respectively.

\noindent\textbf{Step \textcircled{2}.} A and B train each model with locally resided feature(s) to get the intermediate value, in particular, the softmax after the forward computation. Note that B has no label to perform backward propagation so far, and B thus encrypts the softmax with $k_p$ and sends it to A. Then A computes the loss over the encrypted softmax for B (\emph{i.e.}, line 4 in Algorithm~\ref{nowith}). At the same time, A performs forward computation to gain its own softmax and then computes the loss---this is in plaintext. A then computes the aggregated loss in ciphertext by combining all losses and sends the aggregated loss to B.

\noindent\textbf{Step \textcircled{3}.} A and B perform backpropagation over the ciphertext guided by the aggregated loss to update the gradient for their local models. A and B each generate a random number $\mathcal{R}_A$ and $\mathcal{R}_B$ serving as a mask, respectively, and encrypts it firstly with $k_p$. Each A and B sums its encrypted gradient and the encrypted mask and sends the summation to C.

\noindent\textbf{Step \textcircled{4}.} With the private key $k_s$, C decrypts the masked gradient of A and B, and sends each back to A and B. Notably, [[.]] satisfies additive homomorphic~\cite{yang2019federated}. Also, note C cannot gain the plaintext of A or B's gradient because of the usage of the unknown mask to C. Note A/B has added a mask (\emph{i.e.}, a random number) to the gradient, thus C cannot obtain the real value.

\noindent\textbf{Step \textcircled{5}.} For A and B, each subtracts the mask and gains the undated local model parameters in plaintext, which can enter into the next round by repeating Step \textcircled{2} to Step \textcircled{5} till the convergence of the model or a predefined condition is satisfied.

\begin{algorithm}[!t]

\caption{Non-split VFL with coordinator.}
\label{nowith}
\begin{algorithmic}[1]

\REQUIRE Aligned datasets and labels $\mathcal{D}_A$, $\mathcal{D}_B$, $\mathcal{Y}$.
\ENSURE Updated model parameters $\theta_A$ and $\theta_B$.

\STATE Party A and party B initialize parameters $\theta_A$ and $\theta_B$, respectively, and coordinator C creates an encryption key pair and sends the public key to A and B.
\FOR{each iteration}

\STATE Party A calculates softmax value $p_A$; party B calculates softmax value $p_B$ and encrypts it then sends [[$p_B$]]to A.

\STATE Party A calculates loss $l$=$\mathcal{L}$($p_A$+[[$p_B$]]-$\mathcal{Y}$) and encrypts it, then sends [[$l$]] to B.

\STATE Party A and B calculate and encrypt the gradient [[$g_A$]] and [[$g_B$]] according to loss and then generate a random number $\mathcal{R}_A$ and $\mathcal{R}_B$ to mask $g_A$ and $g_B$ and send the [[$g_A$]]+[[$\mathcal{R}_A$]] and [[$g_B$]]+[[$\mathcal{R}_B$]] to C.

\STATE C decrypts [[$g_A$]]+[[$\mathcal{R}_A$]] and [[$g_B$]]+[[$\mathcal{R}_B$]], then sends $g_A$+$\mathcal{R}_A$ and $g_B$+$\mathcal{R}_B$ to A and B respectively. 

\STATE A and B remove the mask, then update the model parameters.

\ENDFOR
\end{algorithmic}
\end{algorithm}

\subsubsection{Non-split VFL without coordinator}

 \begin{figure} [t]
    \centering
    \includegraphics[width=0.3\textwidth]{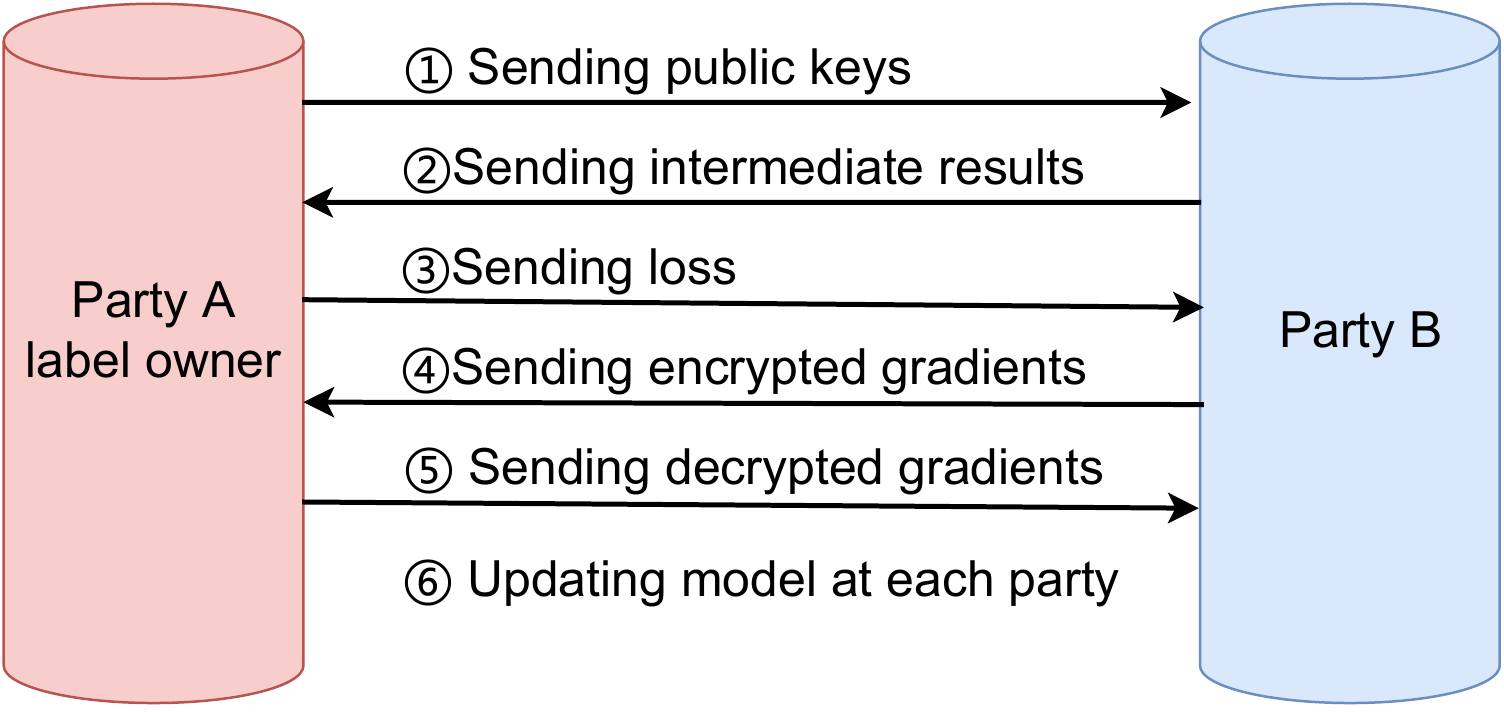}
    \caption{Non-split VFL without a coordinator.}
    \label{fig:without coordinator}
\end{figure}

Figure~\ref{fig:without coordinator} provides the overview of non-split VFL without a coordinator. It has only two types of participants: active party A and passive party B. Algorithm \ref{nowithout}
shows the training process. The training process can be divided into five steps.

\noindent\textbf{Step \textcircled{1}.} A creates a key pair ($(k_p^{(A)},k_s^{(A)})$ and sends the public key $k_p^{(A)}$ to B. A and B initialize their local models, respectively.

\noindent\textbf{Step \textcircled{2}.} A and B train each model with a locally resided feature to get the intermediate value, in particular, the softmax after the forward computation. Because B does not yet have labels to perform back propagation, so B encrypts the softmax with $k_p^{(A)}$ and sends it to A. A computes the loss over the encrypted softmax for B. At the same time, A performs forward computation to gain its own softmax and then computes the loss---this is in plaintext. Then, A calculates the aggregated loss by combining all losses and encrypts the aggregated loss with $k_p^{(A)}$ and sends it to B.  

\noindent\textbf{Step \textcircled{3}.} A performs backpropagation using the aggregated loss to update the gradients of its local model. B also performs backpropagation over the ciphertext guided by the aggregated loss to update the gradient of its local model. B generates a random number $\mathcal{R}_B$ serving as a mask and encrypts it firstly with $k_p^{(A)}$. Then B sums its encrypted gradient with an encrypted mask and sends the summation to A.

\noindent\textbf{Step \textcircled{4}.} With the private key $k_s^{(A)}$, A decrypts the gradient of B that is still under the mask protection and sends it back to B. Note A cannot gain the plaintext of B's gradient because of the usage of the unknown mask to A.

\noindent\textbf{Step \textcircled{5}.} B subtracts the mask and gains the updated local model parameters in plaintext. As for A, it decrypts the ciphertext gradient with the private key $k_s^{(A)}$. Now both can enter into the next round by repeating Step \textcircled{2} to Step \textcircled{5} till the convergence of the model or a predefined condition is satisfied.

The key difference between the non-split VFL without a coordinator and with a coordinator is that in the former, the label owner undertakes the work of the coordinator.

\noindent{\bf Communication and Computation Overhead.} There are several rounds of interactions in the training of the non-split VFL---the initial public key distribution is not counted in the following. More specifically, when the third coordinator is involved, the active participant and the passive participant need to communicate twice; the active participant and the coordinator need to communicate twice, and the passive participant and the coordinator need to communicate twice.
When the third coordinator is not used, the active participant and the passive participant need to communicate four times in each round of training so that the communication overhead can be reduced compared to non-split VFL with the coordinator.

\noindent{\bf Non-split VFL in Plaintext.} The above non-split VFL incurs heavy computation overhead because the computation (\emph{i.e.}, backward propagation) has to be performed over ciphertext. The encryption may be removed to reduce the computation overhead significantly. In this case, the passive participant and the active participant only need to communicate twice to transmit the softmax value and the loss value, respectively, which can greatly reduce the communication overhead as well. From the privacy perspective, the passive participant only discloses its own softmax value, and the active participant discloses the loss value. Neither the local data nor local model gradients are disclosed.

\begin{algorithm}[!t]

\caption{Non-split VFL without coordinator.}
\label{nowithout}
\begin{algorithmic}[1]

\REQUIRE Aligned datasets and labels $\mathcal{D}_A$, $\mathcal{D}_B$, $\mathcal{Y}$.
\ENSURE Updated model parameters $\theta_A$ and $\theta_A$.

\STATE Party A and party B initialize parameters $\theta_A$ and $\theta_B$, respectively, A also creates an encryption key pair and sends the public key to B.
\FOR{each iteration}

\STATE Party A calculates softmax value $p_A$; party B calculates softmax value $p_B$ and encrypts then sends [[$p_B$]]to A. 

\STATE Party A decrypts [[$p_B$]] to get $p_B$. Then A calculates loss $l$=$\mathcal{L}$($p_A$+$p_B$-$\mathcal{Y}$) and encrypts it, then sends [[$l$]] to B.

\STATE Party A and B calculate and the gradient $g_A$ and [[$g_B$]] according to loss.

\STATE Party B generates a random number $\mathcal{R}_B$ to mask $g_B$ and send [[$g_B$]]+[[$\mathcal{R}_B$]] to A.

\STATE Party A decrypts [[$g_B$]]+[[$\mathcal{R}_B$]], then sends $g_B$+$\mathcal{R}_B$ to B; B remove mask to get gradient. 

\STATE A and B update the model parameters.

\ENDFOR
\end{algorithmic}
\end{algorithm}

\subsection{Split VFL}

\begin{figure}
    \centering
    \includegraphics[width=0.3\textwidth]{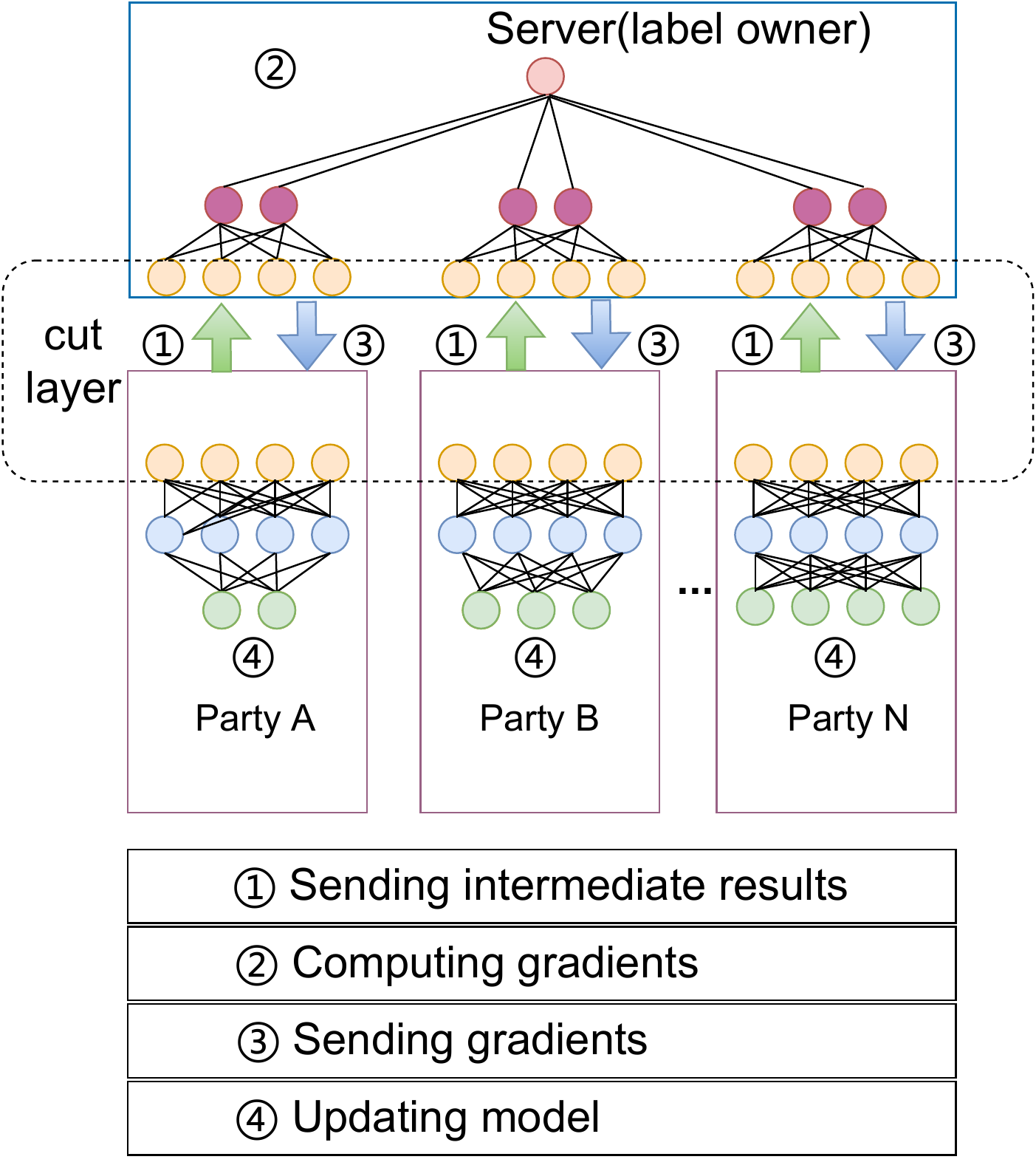}
    \caption{Split VFL.}
    \label{fig:split}
\end{figure}

\begin{figure}
    \centering
    \includegraphics[width=0.3\textwidth]{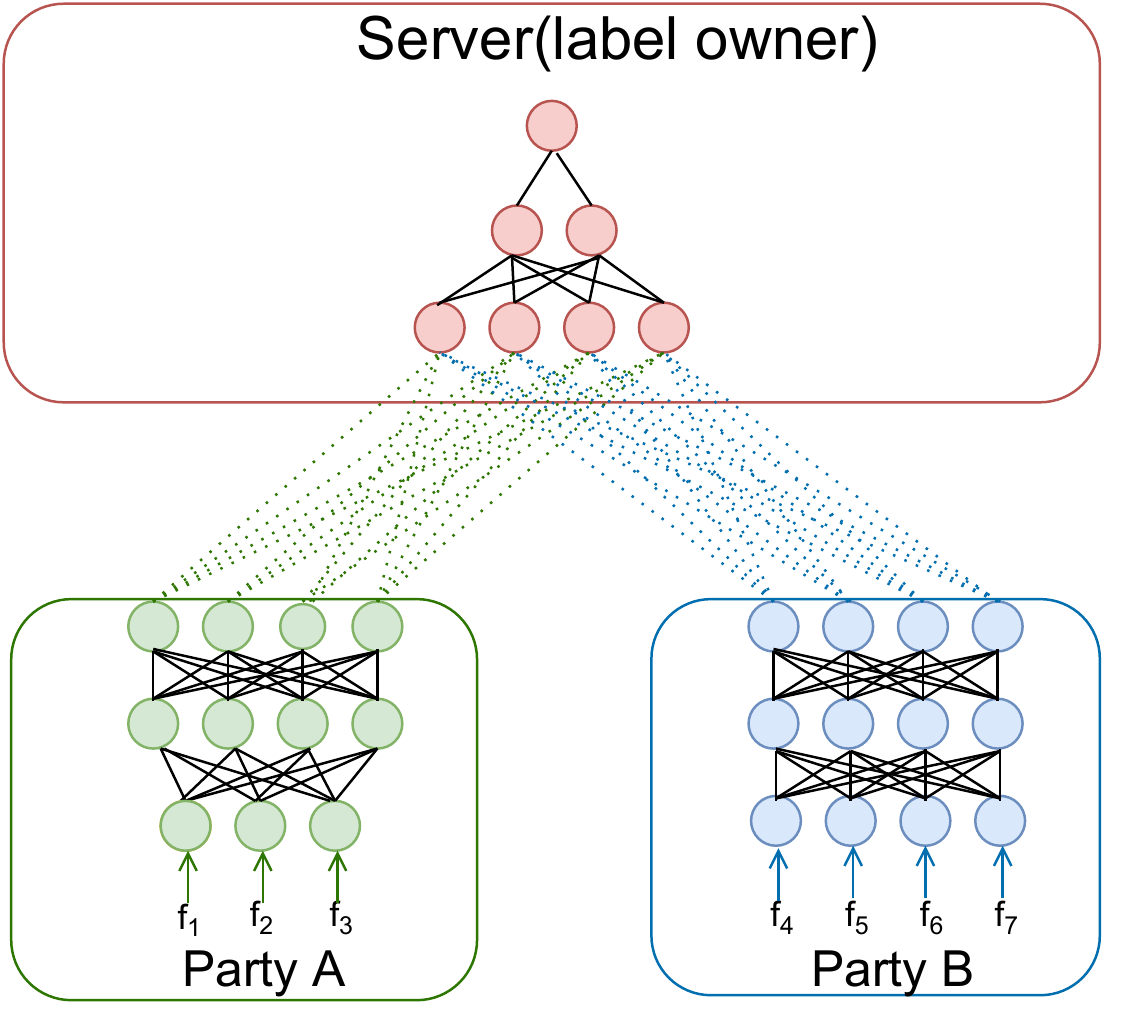}
    \caption{Split neural network.}
    \label{fig:splitnn}
\end{figure}

Figure~\ref{fig:split} shows an overview of the split-based VFL. 
In split VFL~\cite{zhang2021desirable,hashemi2021vertical,kang2020fedmvt}, there are two types of participants: server or active participant (\emph{i.e.}, Party A as host) that holds labels and data; and the rest passive participants (guest) only hold data. Figure~\ref{fig:splitnn} exemplifies a splitted neural network. The entire model is split into a top model and several bottom models. All participants (active and passive) have their own bottom models. The active participant owns the top model. The split occurs at the so-called cut layer, particularly a fully-connected layer. In this example, party A and B has three and four features, respectively, and each has a partial network. The split-based VFL training process can be divided into five steps:

\noindent\textbf{Step \textcircled{1}.} Each participant initializes (\emph{i.e.}, randomize the values of the weights) his/her partial bottom model respectively.

\noindent\textbf{Step \textcircled{2}.} Based on local data, each participant performs forward propagation to gain the intermediate result of the local bottom network, namely smash data, and sends it to the server.

\noindent\textbf{Step \textcircled{3}.} The server collects the smash data from all guests and himself/herself and continues the forward propagation to gain the loss, and then performs the backward propagation that computes the gradients to the cut layer.

\noindent\textbf{Step \textcircled{4}.} The server passes the corresponding gradient to each guest.

\noindent\textbf{Step \textcircled{5}.} The guest, as well as the server, continue the gradient backward propagation to update the bottom model parameters. Step \textcircled{2} to Step \textcircled{5} are repeated till the convergence of the model or a predefined condition is satisfied.

Algorithm~\ref{split} details the steps of the split-based VFL. where ${\bf x}_{n,m}$ is the $m$-th feature of $n$-th sample vector. We note that these split-based VFL studies split-based VFL~\cite{zhang2021desirable,hashemi2021vertical,kang2020fedmvt} are on plaintext setting potentially due to unbearable computation cost of the frequent forward and backward propagation per epoch if they are performed over ciphertext.

\begin{algorithm}[!t]

\caption{Split VFL.}
\label{split}
\begin{algorithmic}[1]

\REQUIRE Aligned datasets $\mathcal{D}_m$, $m \in \{1, ..., \mathcal{M}\}$ and labels $\mathcal{Y}$.

\ENSURE Updated model parameters $\theta_m$.

\STATE Each guest initializes parameters $\theta_m$, server initializes parameters $\theta_0$.
\FOR{each iteration}

\STATE Each guest $m$ uses $\theta_m$ maps the high dimensional vector $x_{n,m}$ to a low-dimensional one $h_{n,m}$ and sends it to server.

\STATE Server collects all $h_{n,m}$ from each guest $m$ and calculates loss $l(\theta_0,h_{n,1},...,h_{n,\mathcal{M}};\mathcal{Y}$) and
gradient $\nabla_{\theta_0}l(\theta_0,h_{n,1},...,h_{n,\mathcal{M}};\mathcal{Y})$.

\STATE Server updates $\theta_0$, then sends gradient
to each guest $m$.

\STATE Each guest $m$ updates their parameters $\theta_m$ with gradient.

\ENDFOR
\end{algorithmic}
\end{algorithm}

\subsection{Customized VFL}

\begin{figure}
    \centering
    \includegraphics[width=0.3\textwidth]{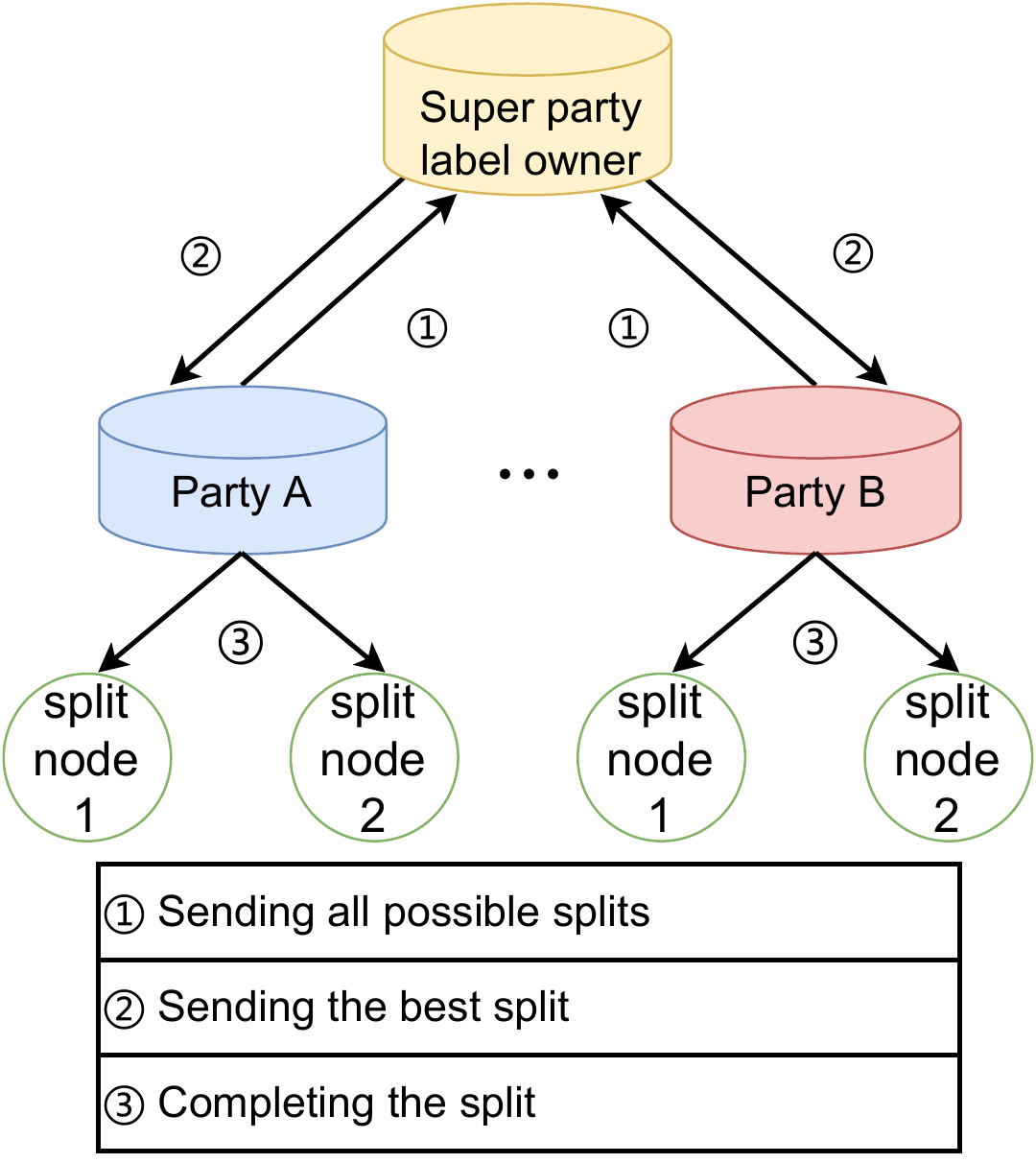}
    \caption{Tree-based VFL.}
    \label{fig:tree}
\end{figure}

There exist customized VFL algorithms that are tree-based~\cite{wu2020privacy,luo2021feature,tian2020federboost}. This algorithm involves a super participant who holds the label and other passive participants who hold different features to jointly train the best classification tree model. The tree-based VFL algorithm is summarized in Algorithm~\ref{tree} and illustrated in Figure~\ref{fig:tree}. Its training process can be divided into three steps:

\noindent\textbf{Step \textcircled{1}.} Each party uses its own local data and features to list all possible splits and send them to the super party.

\noindent\textbf{Step \textcircled{2}.} The super party uses the label to calculate the best split corresponding to each party and then sends it to the corresponding party.

\noindent\textbf{Step \textcircled{3}.} Parties are split according to the best split. Repeat Step \textcircled{1} to Step \textcircled{3} till a complete decision tree model is built.

\begin{figure}
    \centering
    \includegraphics[width=0.3\textwidth]{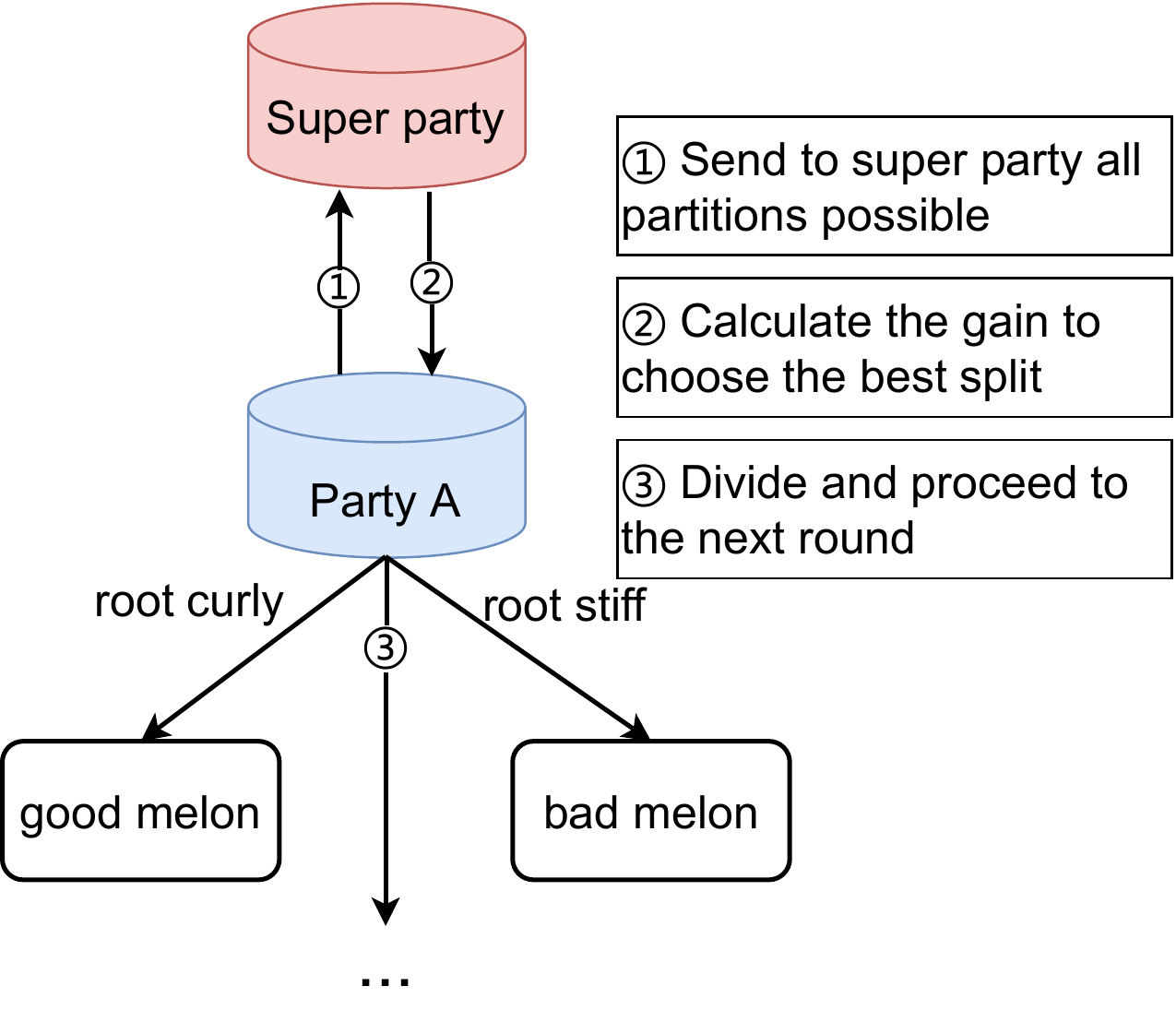}
    \caption{Decision tree split process.}
    \label{fig:treeexample}
\end{figure}

Figure~\ref{fig:treeexample} shows the decision tree splitting process of good or bad melon. First, party A lists all possible splits according to the data features it has (for example, `color', `root', `texture'), and then sends it to the super party. Then the super party calculates the information gain of each split according to its own label, and selects the one with the largest gain as the best split, such as `root'. Return the best split to party A. Party A uses the `root' as the split condition, and the `root' is curled as a good melon, and the `root' is stiff as a bad melon. Then party A continues the split of other features until the decision tree is constructed.

\begin{algorithm}[!t]

\caption{Tree-based VFL.}
\label{tree}
\begin{algorithmic}[1]

\REQUIRE Local datasets {$\mathcal{D}_m$}, $m \in \{1,...,\mathcal{M}\}$, local features {$\mathcal{F}_m$}, and label $\mathcal{Y}$.

\ENSURE Decision tree model.

\FOR{every split when building the tree}

\STATE Party $m$ lists all split modes according to the $\mathcal{F}_m$ it owns and sends it to the super party.

\STATE Super party computes the best split for each party using its own label and sends it to the corresponding party.

\STATE Each party splits according to the best split received.

\ENDFOR

\STATE Repeat the splitting process until a complete decision tree model is built.

\end{algorithmic}
\end{algorithm}



\subsection{Application Scenarios}

\begin{table*}[]
    \centering
    \caption{Application scenarios of different VFL algorithms.}
    \scalebox{0.95}{
    \begin{tabular}{c|c}
    \hline
       \textbf{VFL Algorithms} & \textbf{Application scenarios}\\
    \hline
    
    Non-split VFL with coordinator & Node classification~\cite{ni2021vertical}, medical imaging~\cite{dou2021federated,yang2021federated}, risk management~\cite{ou2020homomorphic}\\
    \hline
    Non-split VFL without coordinator & Vision-and-language grounding problems~\cite{liu2020federated,huang2020federated}\\
    \hline
    Split VFL & Anomaly detection\cite{stolpe2013anomaly}, recommendation system~\cite{wu2022practical} \\
    \hline
   Tree-based VFL & Classification\cite{wu2020privacy}\\
     \hline
     
    \end{tabular}
    }
    \label{tab:application scenarios}
\end{table*}

The success of VFL is mainly attributed to its application scenarios~\cite{fu2021vf2boost}. Compared with independent training or HFL, VFL can take advantage of more/deeper attribute dimensions and obtain better-learned models. VFL has been successfully applied to Fedlearner$\footnote{https://github.com/bytedance/fedlearner}$
in ByteDance, Angel PowerFL$\footnote{https://data.qq.com/powerf}$ in Tencent, FATE$\footnote{https://github.com/FederatedAI/FATE}$ in Webank, Fedlearn$\footnote{https://github.com/fedlearnAI/fedlearn}$ in JD and Paddle$\footnote{https://github.com/PaddlePaddle/PaddleFL}$ in Baidu. We discuss the application scenarios of VFL in the following, and the summary is depicted in Table~\ref{tab:application scenarios}.

Graph machine learning (GML) can effectively process graph data; it has gained increased attention in both academia and industry~\cite{fu2022federated}. With excellent performance, it is widely used in node classification~\cite{zhao2021graphsmote,hang2021collective}, relationship prediction~\cite{cai2021line,daza2021inductive}. Ni {\it et al.}~\cite{ni2021vertical} applied graph neural network to non-split VFL with a coordinator. In this work, participants use a graph neural network as a local model for training and finally train a more functional model. 

Because VFL is suitable for scenarios with the same sample space but different feature space, it is suitable for the field of medical imaging~\cite{nguyen2022federated}. Dou {\it et al.}~\cite{dou2021federated} used the non-split VFL model with coordinator to detect abnormal Computed Tomography (CT) scans in COVID-19 lungs. Yang {\it et al.}~\cite{yang2021federated} proposed a semi-supervised learning framework combined with non-split VFL with coordinator to segment the affected regions of COVID-19 in 3D chest CT. 

Machine learning can quickly predict credit risk, and it is often more accurate than manual prediction. Ou {\it et al.}~\cite{ou2020homomorphic} combined with homomorphic encryption to construct a non-split VFL with coordinator framework compatible with other Bayesian systems for risk management.
 
In recent years, vision-and-language grounding problems, \emph{\emph{e.g.}}, image captioning, and visual question answering (VQA) have aroused widespread concern in academia and industry~\cite{antol2015vqa,goyal2017making}. VFL can also be used to deal with vision-related problems. Liu {\it et al.}~\cite{liu2020federated} proposed a non-split VFL scheme without coordinator to extract fine-grained image representation. Huang {\it et al.}~\cite{huang2020federated} combined the encoder with non-split VFL without coordinator to execute the spoken language understanding (SLU) task. 
 
Unsupervised learning enables using unlabeled data to train models~\cite{ghahramani2003unsupervised}. Unsupervised learning can be applied for anomaly detection ~\cite{hilas2008application,yang2011hybrid}, recommendation system~\cite{bakshi2014enhancing,chiu2021developing}. Wu {\it et al.}~\cite{wu2022practical} proposed a split VFL scheme for unsupervised learning. The passive participant has no label but only data, so it can use unsupervised learning to extract data features and send them to the active participant for training.

Decision tree~\cite{milanovic2016chaid,tayefi2017application} is one of the most commonly used prediction methods for building data mining models. This method is used for classification~\cite{song2015decision} and other predictive data mining tasks~\cite{sharma2016survey}. The VFL scheme proposed by Wu {\it et al.}~\cite{wu2020privacy} can safely construct decision trees for classification. And the decision tree can be extended to the random forest. The super party first broadcasts some encrypted information and lets the party compute the necessary statistics for encryption locally.
The parties then collectively convert these statistics into confidentially shared values to determine the best split of the current tree node using secure computation.
Finally, the best split of the secret sharing is revealed so that the party can update the model.

\begin{table*}[]
    \centering
    \caption{Comparison of different categories of VFL.}
    \scalebox{0.95}{
    \begin{tabular}{c|c|c|c|c}
    \hline
       \textbf{Category} & \textbf{Applicable Models} & \textbf{Participant types} & \textbf{Advantages} & \textbf{Disadvantages} \\
    \hline
    
      \makecell[l]{Non-split VFL \\  with coordinator }&
     \makecell[c]{Logistic regression\cite{yang2019federated}, \\ Neural network (\emph{\emph{e.g.}} DNN\cite{jin2021cafe}, GCN\cite{ni2021vertical})} &
     \makecell[c]{Active participant\\ Passive participant\\ Coordinator} & \makecell[c]{Protect the privacy\\ of participants} & High communication cost\\
     \hline
     \makecell[l]{Non-split VFL \\without coordinator}&
     \makecell[c]{Logistic regression\cite{castiglia2022flexible} \\ Neural network (\emph{\emph{e.g.}} DNN\cite{zhang2022data}, GCN\cite{cheung2021fedsgc})}&
     \makecell[c]{Active participant\\ Passive participant}&
     Low communication cost &
     \makecell[c]{High privacy risk\\ of the passive participant}\\
     \hline
     \makecell[l]{Split VFL} & Neural network (\emph{\emph{e.g.}} DNN\cite{hashemi2021vertical}, GCN\cite{qiu2022your})&  \makecell[c]{Active participant\\ Passive participant}& 
     \makecell[c]{Protect the privacy\\ of participants}& 
     \makecell[c]{Split original model\\ loss accuracy}\\
     \hline
     \makecell[l]{Tree-based VFL} & Tree model\cite{wu2020privacy,luo2021feature,tian2020federboost} & 
     \makecell[c]{Active participant\\ Passive participant}& 
     Fast inference speed&
     \makecell[c]{Easy over fitting\\ Poor generalization}\\
     \hline
     
    \end{tabular}
    }
    \label{tab:comapre}
\end{table*}

\section{Security and Privacy Threats}\label{sec:SecPri}
The threats to VFL can be broadly categorized into security attacks and privacy inference attacks.

\subsection{Security Threats}

Security attacks demonstrated against VFL include backdoor attacks introduced during the training phase to compromise the integrity of the model~\cite{liu2021defending}, and adversarial example attacks occurred solely during the inference phase~\cite{chen2022graph}. The active participant and passive participants have \textit{different} levels of threats to the VFL due to their differing knowledge and control capabilities over the VFL. The VFL can be threatened by poisoning attack introduced during the training phase to compromise the models' availability (\emph{i.e.}, deteriorating the overall model accuracy) during the inference phase~\cite{gao2020backdoor}, although such attacks have not been explicitly demonstrated against the VFL. 

\subsubsection{Backdoor Attacks} 
The attacker inserts a backdoor into the DL model that causes the DL model to misbehave only when a secret trigger is present in the input~\cite{gao2020backdoor,gao2021evaluation}. For example, an infected face-recognition model can still correctly classify Alice's face image to be Alice. But when a pair of black-frame glasses is worn, the model wrongly classifies Alice's face image as an administrator.

The most common backdoor is implanted through data poisoning in the model outsourcing scenario~\cite{gu2019badnets}. The attacker randomly selects a small fraction of samples, stamping the trigger on each of them, changing their label to the target label, and then mixing the poisoned samples with clean samples to train the model, which is backdoored.

Such a data-poisoning-based backdoor is straightforward to the active participant in the VFL. As the active participant holds the label and data, the label can be readily manipulated to generate poisoned data. However, such a data-poisoning-based backdoor is not immediately applicable to passive participants as he/she cannot modify the label.

Liu {\it et al.}~\cite{liu2020backdoor} proposed a gradient replacement backdoor attack suitable for passive actors in non-split VFL. Specifically, it assumes that the attacker is a passive participant, and it knows at least one clean sample, say $\mathcal{D}_{\rm Target}$, from the target label of the backdoor attack. 
The attacker records the received intermediate gradient $g_{\rm rec}$ for  $\mathcal{D}_{\rm Target}$. It then replaces the intermediate gradient of its poisoned sample to $g_{\rm rec}$ and uploads the replaced gradient in each round of interaction with the active participant to update the model. 

Qiu \textit{et al.}~\cite{qiu2022hijack} proposed to hijack or flip the VFL's prediction result (\emph{e.g.}, from  `not qualified' to `pass' for a loan application) by a passive participant who has no label information and only a small fraction of controlled features. The key is to use robust features such as income that dominantly determine the target class (\emph{e.g.}, `pass' a loan application) to replace the passive participant local features. The robust features can be searched from existing features or even synthesized through optimization. To make the robust features more influential on the target class prediction, a poisoning attack is also leveraged during the VFL training phase to enforce the VFL model to learn a strong connection between the (adversarial) robust feature and the target class.

\subsubsection{Adversarial Example Attacks}

The adversarial example (AE) attack is crafted by delicately adding a small disturbance that cannot be recognized by the human eye to the input sample fed into the model. So that the adversarial example (\emph{i.e.}, adversarial `dog' image) is misclassified (\emph{i.e.}, into `cat' class) by the target model with high confidence~\cite{qiu2019review} even though the AE and its underlying original sample (\emph{i.e.}, a clean `dog' image) are same to the human.

Chen {\it et al.} proposed an AE attack against graph VFL in the inference phase, namely graph fraudster~\cite{chen2022graph}. Firstly, a malicious participant is randomly selected from all passive participants. A malicious participant uses the generative regression network (GRN) to steal the intermediate information (\emph{i.e.}, the node embeddings extracted by the local graph neural network) uploaded by other participants and build a shadow model of the server for the attack generator. Then, noise is added to the node embeddings to attack the shadow model through iterative adversarial perturbation optimization. The produced AE on the shadow model can also then be transferred to attack the server model.

\subsection{Privacy Threats}

Despite FL being resistant to privacy leakage as it avoids direct sharing of raw data required for centralized training, it is still vulnerable to various privacy inference attacks~\cite{zhou2022ppa}. 
The VFL has been shown to be vulnerable to label inference attacks, data reconstruction attacks, and attribute inference attacks, each of which is detailed in the following.

\subsubsection{Label Inference Attacks} In VFL, only active participants have labels, which is different from HFL in which each participant has its own labeled sample. The label should be protected because it could be valuable assets of active participants or highly sensitive assets~\cite{fu2022label,sun2022label}. For example, the institution that provides labels can be a hospital, and these labels are information on whether a patient has a certain disease. The gradients propagated during training by VFL often help the attacker's local model to learn feature representations about labels so that the local model may contain information about the labels. Building upon such rationale, Fu {\it et al.}~\cite{fu2022label} proposed three label inference attacks against VFL, including active label inference attack and passive label inference attack against split VFL, and direct label inference attack against non-split VFL. Note such attacks occur when the VFL is performed in plaintext.

In the passive label inference attack, the attacker accesses a small number of active participant features and corresponding labels, namely auxiliary labeled data. The attacker adds an additional random initialization layer on top of her bottom model in the split VFL to construct a full model for label inference. Assisted by the auxiliary data, the attacker can use semi-supervised learning to fine-tune the trained bottom model to obtain a complete label inference model to launch label inference. Because the attacker keeps being honest but curious, \emph{i.e.} she strictly follows the VFL protocol in both the training and inference phases. More specifically, the attacker will execute all the processes of the VFL protocol, including the training process and inference process, but will generate an additional network layer in its own local model. They are combining additional layers with their own model, using auxiliary labels for inference. This attack is thus a passive attack.

In the active label inference attack, the attacker uses a specially designed local optimizer instead of common optimizers such as Adam~\cite{kingma2014adam} or stochastic gradient descent (SGD). The attacker can thus accelerate gradient descent on her own bottom model, learning a better bottom model. It ultimately makes active participants more dependent on the bottom model controlled by the attacker than other participants. Then the attacker can obtain the complete label inference model according to the method in the passive inference attack.
Overall, the attacker lures the active participant to leak more label-related information by delicately interacting with the active participant to update the attacker's local model.

In a direct label inference attack, the attacker infers labels by analyzing gradients in plaintext transmitted during non-split VFL training. Since the loss is computed using the error between the predicted value and the true label, the gradient information returned by the server leaking the label information. In this attack, Fu {\it et al.}~\cite{fu2022label} assumed that the passive participant in a non-split VFL can gain the gradient returned by the active participant. Similar to the HFL server aggregating the outputs of all passive participants to compute the loss and gradient, the non-split VFL active server returns the gradient to all passive participants. 

Zou \textit{et al.}~\cite{zou2022defending} proposed a label replacement backdoor attack against non-split VFL under ciphertext conditions. The attacker, which is a passive participant, has access to some clean samples with the target label, denoted as $\mathcal{D}_t$, and the poisoned samples are denoted as $\mathcal{D}_\textup{p'}$. During the forward propagation of each training round, if the sample $i$ belongs to $\mathcal{D}_\textup{p'}$, then the attacker randomly selects a sample $j$ from $\mathcal{D}_t$, replaces the encrypted intermediate value [[$p_B = h_i$]] with [[$p_B=h_j$]]. The attacker then sends $h_j$ to the active party. In backward propagation, the attacker replaces the received intermediate gradient of a sample $i$ with the encrypted gradient of $j$ and updates the local model. Note the attacker is the passive participant who knows $\mathcal{R}_B$. By using this strategy, the passive party will get the corresponding gradient against the target label instead of its own, so its local backdoor update will be successful.

\subsubsection{Data Reconstruction Attacks} In a data reconstruction attack, the attacker reverse-engineers the training data of the participant to breach the data privacy. Jiang {\it et al.}~\cite{jiang2022comprehensive} proposed an attack leveraging confidence score to reconstruct data in the inference phase of the non-split VFL.
The attacker has access to a small amount of auxiliary data, that is, the passive participant's partial data $\mathcal{D}_{\rm pas}$ and the corresponding confidence score $\mathcal{C}_{\rm pas}$. The attacker first initializes a shadow model and then uses the auxiliary data to calculate the confidence score $\mathcal{C}_{\rm act}$ to compute the distance between $\mathcal{C}_{\rm pas}$ and $\mathcal{C}_{\rm act}$, and continuously minimize the difference between $\mathcal{C}_{\rm pas}$ and $\mathcal{C}_{\rm act}$ to  approach the functional behavior of the true passive participant model. 
After obtaining the optimized shadow model, the attacker initializes an estimated passive participant data, then computes the distance between the shadow model's confidence score and the true confidence, and minimizes this distance to recover the passive participant's data.

Jin {\it et al.}~\cite{jin2021cafe} proposed a massive data disclosure attack scheme against non-split VFL with the coordinator. Suppose the attacker is a coordinator and knows the index of the data. The data index is the ID list of the data selected for each training. Before each training, the attacker selects the data index, and the participants select their partial feature data for training according to the index. The attacker initializes the loss, then iteratively computes the loss according to the gradient sent by the participant. According to the loss, the attack consequentially reconstructs the input according to the loss, that is, the feature representation extracted from the data. Finally, the attacker restores the original data according to the input.

Weng {\it et al.}~\cite{weng2020privacy} proposed a reverse multiplication attack against non-split VFL gradient encryption in the inference phase to infer the target participant's original training data. In this attack, coordinator C is assumed to have already broken or colluded with the attacker, the attacker A has thus obtained the private key and the gradient of the passive participant B---thus essentially equal to a plaintext attack. Then the attacker can obtain the decrypted intermediate results of two consecutive rounds $\theta_{t}^{B}\mathcal{X}^{B}$, $\theta_{t-1}^{B}\mathcal{X}^{B}$ and gradient $g_{t}^{B}$ of the participant B. Note that intermediate results $\theta_{t}^{B}\mathcal{X}^{B}$ is obtained/known as a whole, where $\theta$ and $\mathcal{X}$ are simply two arguments to compute the intermediate result---with $\theta$ the coefficient vector for each round of gradient calculation. According to the equation,

\begin{equation}
    \theta_{t}^{B}\mathcal{X}^{B}- \theta_{t-1}^{B}\mathcal{X}^{B}=g_{t}^{B}\mathcal{X}^{B},
\end{equation}
the attacker knows all parameters except the original data of the passive participant $\mathcal{X}^{B}$ and then can solve this equation to gain the original data $\mathcal{X}^{B}$.

\subsubsection{Attribute Inference Attacks}
Attribute inference attacks aim to infer sensitive attributes of samples~\cite{yeom2018privacy}. In VFL, both the active participant and the passive participant have their own data/feature. The data contain attribute information, such as gender and age, which can be sensitive. The attacker attempts to obtain the privacy attributes of the participants. The attribute inference attack on the VFL can occur in the training phase or the inference phase.

\vspace{2pt}\noindent$\bullet \hskip3pt $\textbf{In Training Phase.} 
Zhang {\it et al.}~\cite{zhang2021privacy} assumed that the attacker has partial passive participant's data $D_m$ and corresponding intermediate outputs (\emph{e.g.}, softmax) $\mathcal{O}_m$. The attacker's goal is to train a decoder represented by its parameter $w_m$ on ($\mathcal{O}_m, \mathcal{D}_m$) by Eq.~\ref{inference_attack}. Once trained well, the decoder is able to reconstruct input features when it inputs the intermediate output.
\begin{equation}
    \min_{w_m} l(w_m;\mathcal{O}_m,\mathcal{D}_m)
    \label{inference_attack}
\end{equation}

\begin{figure}
    \centering
    \includegraphics[width=0.3\textwidth]{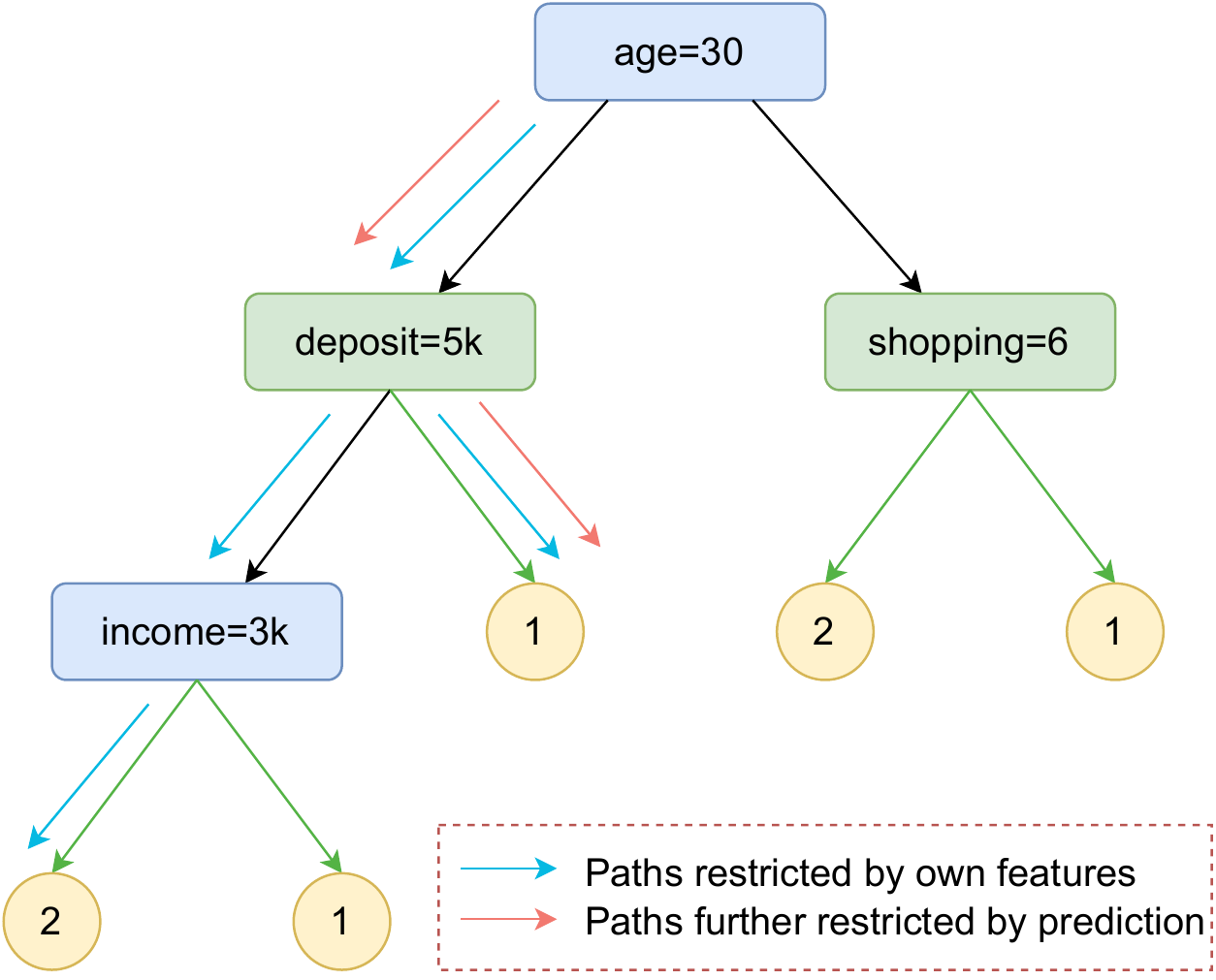}
    \caption{Example of path restriction attack.}
    \label{fig:attacktree}
\end{figure}

\vspace{2pt}\noindent$\bullet \hskip3pt $\textbf{In Inference Phase.} Luo {\it et al.}~\cite{luo2021feature} proposed an attribute inference attack against tree VFL.
The active participant uses its own features to compare with the branch/node threshold of the tree.  The active participant can limit the possible predicted paths in the tree model according to his/her owned features. The attacker can determine predicted paths based on predicted labels and consequentially infers the attributes owned by the passive participant. We take an exemplification in Figure~\ref{fig:attacktree}, assuming that the attacker has features `income=2k' and `age=25', then the decision-making path can be limited from the original possible five paths (with green arrows) to two possible paths (with blue arrows). If the passive participant provides a feature for prediction, it can only make predictions through the path indicated by the blue arrows. If the prediction label is 1, then the active participant will find the real prediction path (with red arrows), and it can be inferred that the passive participant has the feature that the deposit is greater than 5k.

\subsubsection{Relation Inference Attack} The VFL has been adopted to learn graph data~\cite{zhou2020vertically}. The graph is good at modeling complex relationships between entities, which can be private information (\emph{e.g.}, dating, debt, and even sexual relationships). Although the learning can be protected through an encryption mechanism to prevent graph data disclosure to other parties, it has shown that the sample relationship can be deduced by a semi-honest passive participant~\cite{qiu2022your}. The key is that a small distance between two samples' latent representations in the split VFL reveals their close relationship. To circumvent the encryption-protected representations, a numerical approximation method is devised to identify a representation meeting constraints that the approximated representations as input to the trained VFL model can \textit{reproduce} the corresponding prediction results.

\begin{table*}[]
    \centering
    \caption{Differences between HFL and VFL in terms of security and privacy threats.}
    \scalebox{0.95}{
    \begin{tabular}{c|c|c|c}
    \hline
       \textbf{Type} & \textbf{HFL} & \textbf{VFL} & \textbf{Threats}  \\
       \hline
       Data ownership  & Each participant owns features and labels & \begin{tabular}{c}
        Active participant has the label, \\   the passive participant has the features 
       \end{tabular} & \begin{tabular}{c}
        VFL is vulnerable to\\  label inference attacks
       \end{tabular} \\
       \hline
       Data alignment & No need to align data & Need to align data & \begin{tabular}{c}
         VFL is vulnerable to \\ attribute inference attacks \\ and data reconstruction attacks  
       \end{tabular} \\
       \hline
       Inference method& \begin{tabular}{c}
        Participants infer the global model independently  \\ and obtain different inference results
       \end{tabular}& \begin{tabular}{c}
         Participants need to jointly reason  \\ and output the same reasoning result  
       \end{tabular} & \begin{tabular}{c}
        VFL is vulnerable to \\adversarial attacks during\\ the inference stage 
       \end{tabular}\\
     \hline
    \end{tabular}
    }
    \\

    \label{tab:HFLlvsVFL}
\end{table*}

\section{Challenges and Prospects}\label{sec:ChaPro}

In addition to the security and privacy challenges, we further discuss other challenges confronting VFL and prospect potential solutions.

\subsection{Security and Privacy Risks}

Considering the privacy threats confronting VFL,
the designed VFL scheme needs to protect the privacy of all participants effectively. Existing privacy-preserving techniques, including Differential Privacy (DP)~\cite{geyer2017differentially,mugunthan2020privacyfl}, Secure Multi-Party Computation (SMC), Homomorphic Encryption (HE)~\cite{zhang2020batchcrypt,ma2022privacy,xu2019hybridalpha}, and their hybrids, have been extensively adopted in HFL but much less explored in VFL, which are worth exploitation. Nonetheless, these methods require carefully balanced trade-offs in model performance (\emph{i.e.}, accuracy), privacy, and system efficiency (\emph{i.e.}, especially for participants with limited resources).
The trade-off evaluations are attracting increasing attention and are being explored. For example, Kang {\it et al.}~\cite{kang2022framework} propose an evaluation framework that formulates the trade-off between privacy leakage and utility loss for evaluating VFL systems. These assessments can facilitate VFL practitioners in choosing an appropriate protection mechanism and tuning corresponding protection parameters given their customized requirements.

One interesting and unique issue of VFL is the private set intersection (PSI)~\cite{liu2020asymmetrical,lu2020multi} used to align IDs of all samples while preserving security and privacy, which has wide applications in decision tree learning, and Naive Bayes classification. In many real-world privacy-sensitive organizations, member information is highly sensitive and cannot be shared with other parties. For example, some participants are companies with high privacy protection requirements on the ID, such as loan companies, and others are companies that do relatively less care about privacy, such as banks. 
After PSI, the IDs of the loan companies will be disclosed to the banks, and the bank can obtain sensitive private information about who has made loans in the company. A trusted third party can be introduced to complete PSI~\cite{kerschbaum2012outsourced}. Using oblivious transfer (OT) to accomplish PSI is also a viable solution~\cite{pinkas2014faster,pinkas2018scalable}. However, designing an efficient PSI protocol that effectively protects privacy is still challenging. 

In fact, there is always a dilemma between security and privacy. Providing privacy protections through cryptographic means inevitably renders the hardness of detecting/preventing security attacks~\cite{ma2022mud} because existing security attack (\emph{e.g.}, backdoor attack, and poisoning attack) countermeasures can usually be applied in plaintext, whereas ciphertext based countermeasures require considerable exploration in the future. 

\subsection{High Computational and Communication Overhead} Similar to HFL, VFL also has the problem of incurring high communication and computational overhead~\cite{singh2019detailed}. The computational overhead in VFL is proportional to the number of participants and samples. Taking the VFL with coordinator in Fig.~\ref{fig:with coordinator} as an example, assuming that the public key has been sent, during each round of training, A and B need to communicate twice, A and C, and B and C need to communicate twice, respectively. In order to ensure privacy, the data transmitted by communication generally use ciphertext after homomorphic encryption. Therefore, the communication overhead will be very large. Computation over the ciphertext is also substantially increased compared to the plaintext-based computation. Effectively reducing the computational and communication overhead is a major challenge when designing a VFL scheme.

Wu {\it et al.} proposed FedOnce~\cite{wu2022practical}, a one-shot VFL protocol. It allows the passive participants to use unsupervised training locally to extract features and then send them to the active participant to train the model. The entire training process only requires the passive participants to communicate with the active participant once, greatly reducing communication overhead. In addition to this work, to reduce computational and communication costs, three improvements can be considered: 1) pruning the neural network, 2) reducing the size of transmitted messages through compression (\emph{i.e.}, quantization), and 3) choosing an appropriate attribute splitting design.

\subsection{Structural Damage of Model Splitting} 

The splitting of the model at the cut layer of the split VFL (see Figure~\ref{fig:splitnn}) will inevitably cause some performance deterioration to the model performance~\cite{wei2022vertical}. Some original connections in the neural network will be disconnected due to splitting, which can damage the structure to some extent. Therefore, the splitting criterion is imperative in the split VFL scheme to mitigate performance degradation.

\subsection{System and Data Heterogeneity} In VFL, asynchronous updates can occur due to heterogeneous computation resource (\emph{i.e.}, CPU or GPU) and communication bandwidth (\emph{i.e.}, wired or wireless networks) available to each participant~\cite{chen2020vafl,gu2021privacy}. For example, if a participant has a large number of attributes, but its own resource for local training is restricted by memory size, computation capability, and further limited bandwidth for communication, an asynchronous high latency will be introduced. Therefore, the VFL scheme should be devised to tolerate asynchronous updates of participants. 
In addition to the above system heterogeneity, some participants' data features/attributes may contribute negligibly or even adversely to the trained model performance. Therefore, it is important to pre-identify those participants to weigh less on their contribution or even neglect their contribution.

\subsection{Model Fairness} Due to the training data imbalance or bias, its trained model may result in biased prediction~\cite{mehrabi2021survey}. For example, the gender prediction of nurses is biased toward females. Therefore, to prevent some sensitive bias from unfairly affecting the model's prediction, research on model fairness has arisen~\cite{caton2020fairness,kleinberg2018algorithmic}. In order to ensure the fairness of the model, most existing fair ML methods rely on the centralized storage of fair sensitive features to achieve model fairness. However, in VFL, features are distributed to different isolated participants, which makes it difficult to directly apply existing fair ML methods to improve the fairness of VFL models. Second, in order to train a fair model, it is necessary to force each participant to complete the update in each communication round, which is challenged by the system heterogeneity per participant. 
Liu {\it et al.} proposed a method to ensure model fairness in VFL~\cite{liu2021achieving}. They define a term called difference of equal opportunities (DEO), which is the absolute value of the distance difference between the two attributes, to measure model fairness. To efficiently address the difficult DEO optimization problem, they use Lagrangian relaxation to transform it into a nonconvex-concave min-max problem and then use asynchronous gradient coordinate descent to solve it.

Qi {\it et al.}~\cite{qi2022fairvfl} proposed a fair VFL framework called FairVFL. It divides features into two groups, fairness-insensitive features for the target task and fairness-sensitive features that are causally irrelevant to model predictions. More specifically, the participants holding fairness-insensitive features first learn their local data representations, which are then uploaded to the trusted third-party server to construct a unified representation. The unified representation is sent to the passive participant who holds the label to gain the global model. Note that the unified representation can still be biased, attributing in-explicit correlations to fairness-sensitive features. Therefore, the unified representation is sent to the participants with fairness-sensitive features to learn adversarial gradients to remove biases encoded in them. To mitigate potential private fairness-insensitive features encoded by the unified representation, contrastive learning is applied to remove the private information.

\section{Conclusion}\label{sec:conclusion}

In this survey, we first combed and classified existing VFL algorithms. Then we introduced privacy and security issues in VFL. Finally, the challenges and solutions of VFL were studied from four aspects: security and privacy risk, computational and communicational cost overhead, structural damage, and system heterogeneity. Through this work, we present an overall picture of VFL enabling and motivating further studies in its open problems.

\balance
\bibliographystyle{IEEEtran}
\bibliography{References}



\end{document}
\endinput